\makeatletter\def\graphicscache@inhibit{true}\makeatother

\documentclass[letterpaper, 10 pt, conference]{ieeeconf}  %

\IEEEoverridecommandlockouts                              %

\usepackage{graphicx} %
\usepackage[utf8]{inputenc}
\usepackage{amsmath,amsfonts,booktabs,cite} %
\usepackage{gensymb} %
\usepackage{siunitx,textcomp} %
\usepackage[hidelinks]{hyperref}
\usepackage{cleveref} %
\usepackage{subcaption}
\usepackage{tabularx}
\usepackage{makecell}
\usepackage{pbox}
\let\labelindent\relax
\usepackage[inline]{enumitem} %
\usepackage[nolist]{acronym} %
\usepackage{multirow}
\usepackage[ruled,linesnumbered, vlined]{algorithm2e}
\usepackage{bm}
\usepackage{titlesec}
\usepackage{balance}
\usepackage[export]{adjustbox}
\usepackage{cellspace}
\usepackage{enumitem}
\usepackage[font=small, justification=justified]{caption}

\def\secref#1{Sec.~\ref{#1}}
\def\figref#1{{Fig.~\ref{#1}}}
\def\tabref#1{{Tab.~\ref{#1}}}
\def\eqref#1{Eq.~(\ref{#1})}

\newcommand\etal{~\emph{et al. }}

\newsavebox{\twosubbox}

\usepackage{tikz}
\usetikzlibrary{arrows}
\usetikzlibrary{positioning,calc}
\usetikzlibrary{decorations.pathreplacing}
\usetikzlibrary{decorations.markings}
\usetikzlibrary{fit}
\usetikzlibrary{shapes.callouts}
\usetikzlibrary{shapes.geometric}
\usetikzlibrary{matrix}
\usetikzlibrary{arrows}
\usetikzlibrary{decorations.pathmorphing}
\usetikzlibrary{calligraphy}
\usetikzlibrary{shapes.arrows}
\usetikzlibrary{spy}
\usepackage{setspace}

\tikzset{
manip/.style={fill=Pastel2-A},
hardware/.style={fill=Pastel2-B},
perception/.style={fill=Pastel2-D},
system/.style={fill=Pastel2-C},
otg/.style={fill=Pastel2-B},
}

\pgfdeclarelayer{background}
\pgfdeclarelayer{foreground}
\pgfsetlayers{background,main,foreground}

\usepackage{scalefnt}

\usepackage{pgfplots}
\pgfplotsset{compat=1.14}
\usepgfplotslibrary{groupplots}
\usepgfplotslibrary{units}
\usepgfplotslibrary{statistics}
\usepgfplotslibrary{colorbrewer}

\usepackage{pgfplotstable}

\crefname{algocf}{alg.}{algs.}
\Crefname{algocf}{Algorithm}{Algorithms}

\titleclass{\subsubsubsection}{straight}[\subsection]

\newcounter{subsubsubsection}[subsubsection]
\renewcommand\thesubsubsubsection{\thesubsubsection.\arabic{subsubsubsection}}

\titleformat{\subsubsubsection}
{\normalfont\normalsize\bfseries}{\thesubsubsubsection}{1em}{}
\titlespacing*{\subsubsubsection}
{0pt}{3.25ex plus 1ex minus .2ex}{1.5ex plus .2ex}

\makeatletter
\renewcommand\paragraph{\@startsection{paragraph}{5}{\z@}%
	{3.25ex \@plus1ex \@minus.2ex}%
	{-1em}%
	{\normalfont\normalsize\bfseries}}
\renewcommand\subparagraph{\@startsection{subparagraph}{6}{\parindent}%
	{3.25ex \@plus1ex \@minus .2ex}%
	{-1em}%
	{\normalfont\normalsize\bfseries}}
\def\toclevel@subsubsubsection{4}
\def\toclevel@paragraph{5}
\def\toclevel@paragraph{6}
\def\l@subsubsubsection{\@dottedtocline{4}{7em}{4em}}
\def\l@paragraph{\@dottedtocline{5}{10em}{5em}}
\def\l@subparagraph{\@dottedtocline{6}{14em}{6em}}
\makeatother

\IfFileExists{graphicscache.sty}{\usepackage[dpi=400]{graphicscache}}

\DeclareRobustCommand{\compb}[2][system]{\strut\tikz[baseline]{\node[draw,#1,anchor=base,rounded corners,minimum height=1.0ex,minimum width=1.5ex,text depth=0pt]{#2};}}

\newsavebox\CBox

\title{\LARGE \bf
HortiBot: An Adaptive Multi-Arm System for\\Robotic Horticulture of Sweet Peppers
}

\makeatletter
\newcommand{\linebreakand}{%
  \end{@IEEEauthorhalign}
  \hfill\mbox{}\par
  \mbox{}\hfill\begin{@IEEEauthorhalign}
}
\makeatother

\author{Christian Lenz$^{*1,3}$\and Rohit Menon$^{*2,3}$\and Michael Schreiber$^1$\and Melvin Paul Jacob$^2$\linebreakand Sven Behnke$^{1,3,4}$\and Maren Bennewitz$^{2,3,4}$%
\thanks{$^1$: Autonomous Intelligent Systems Lab, University of Bonn, Germany}
\thanks{$^2$: Humanoid Robots Lab, University of Bonn, Germany}
\thanks{$^3$: Center for Robotics, Bonn, Germany}
\thanks{$^4$: The Lamarr Institute, Bonn, Germany}
\thanks{This work has partially been funded  by the Deutsche Forschungsgemeinschaft (DFG, German Research Foundation) under Germany’s Excellence Strategy, EXC-2070 -- 390732324 -- PhenoRob.}
  }

\begin{document}

\maketitle
\thispagestyle{empty}
\pagestyle{empty}

\def\thefootnote{*}\footnotetext{These authors contributed equally to this work.}
\renewcommand{\thefootnote}{\arabic{footnote}}

\begin{abstract} Horticultural tasks such as pruning and selective harvesting are labor intensive and horticultural staff are hard to find.
Automating these tasks is challenging due to the semi-structured greenhouse workspaces, changing environmental conditions such as lighting, dense plant growth with many occlusions, and the need for gentle manipulation of non-rigid plant organs.
In this work, we present the three-armed system HortiBot, with two arms for manipulation and a third arm as an articulated head for active perception using stereo cameras.
Its perception system detects not only peppers, but also peduncles and stems in real time, and performs online data association to build a world model of pepper plants.
Collision-aware online trajectory generation allows all three arms to safely track their respective targets for observation, grasping, and cutting.
We integrated perception and manipulation to perform selective harvesting of peppers and evaluated the system in lab experiments.
Using active perception coupled with end-effector force torque sensing for compliant manipulation, HortiBot achieves high success rates in our indoor pepper plant mock-up.

\end{abstract}

\begin{tikzpicture}[remember picture,overlay]
  \node[anchor=north,align=center,font=\sffamily\small,yshift=-0.4cm] at (current page.north) {%
  IEEE/RSJ International Conference on Intelligent Robots and Systems (IROS), Abu Dhabi, UAE, October 2024.};
\end{tikzpicture}%

\section{Introduction}
\label{sec:intro}

Horticultural tasks such as pruning, thinning, pollination, and selective harvesting are labor-intensive and need to be carried out several times a season~\cite{bac2014harvesting}.
In contrast to the mechanization of large-scale grain and cereal farms, the automation of precision horticulture requires robots.
Robotic manipulation in horticulture presents several challenges due to semi-structured greenhouse workspaces, variations in environmental conditions such as lighting, complex and irregular plant structures, varying plant organ sizes and shapes, dense plant growth with many occlusions and obstacles, and the need for gentle manipulation of non-rigid plant organs~\cite{kootstra2021selective}.

While there is an extensive body of work focusing on fruit detection and localization, research on the full robotic harvesting pipeline is limited~\cite{zhou2022intelligent}.
Most selective harvesting systems use specialized hardware for manipulators and end-effectors~\cite{bac2014harvesting}.
In a recent review, Rajendran~\etal~\cite{rajendran2023towards} suggest equipping selective harvesting robots with cooperative active and interactive perception for improved fruit detection and force sensing-enabled two-arm manipulation capabilities---to match humans in handling complex fruit clusters.
With humanoids having potential to become general-purpose autonomous workers adapting to different tasks~\cite{tong2024advancements}, we aim to close the research gap in horticulture manipulation by proposing a non-specialized solution.
HortiBot is a three-arm system for active perception and dual-arm manipulation in horticulture.
The highly flexible robot is built from off-the-shelf components for multiple horticultural tasks.
Unlike most other works that focus on only vision, control, or motion planning, we present a fully integrated system. Our contributions include:
\begin{itemize}[leftmargin=3ex]
	\item work space analysis and design of a three-arm system with stereo cameras and force-torque sensors,
	\item visual perception of sweet pepper plants combining fruit instance mapping with a novel peduncle detection approach and stem detection,
	\item online active perception during manipulation for refining of targeted pepper and peduncle localization,
	\item dual-arm manipulation using parameterized motion primitives and collision-aware online trajectory generation, and
  \item a thorough evaluation of the selective harvesting capabilities in lab experiments using real sweet peppers.
\end{itemize}

\begin{figure}[t]
	\centering
	\includegraphics[width=0.97\columnwidth,clip,trim=200px 0px 00px 200px]{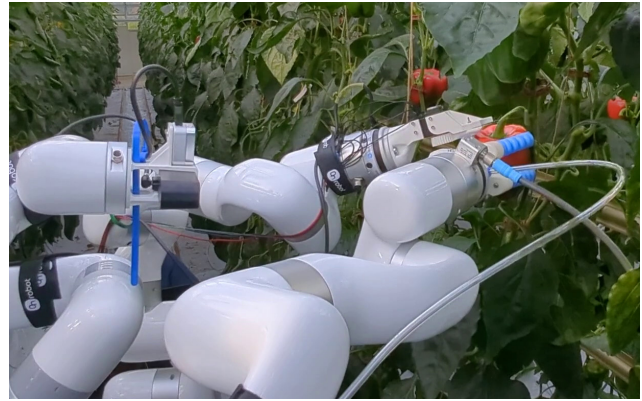}
	\captionsetup{width=0.99\columnwidth, justification=justified}
	\caption{HortiBot: A three-arm system with active perception and dual-arm manipulation for robotic horticulture. The right arm is used for grasping, the left arm performs cutting, and the central arm moves stereo cameras for mapping and online observation.}
	\label{fig:cover}
\end{figure}
 
\section{Related Work}
\label{sec:related}
With advancements in robotics and deep learning methods, different aspects of horticulture have been automated using robotic systems such as pollination~\cite{li2022design} and dormant pruning~\cite{you2023semiautonomous}.
Of the many tasks in the horticultural industry, selective harvesting is the one most often addressed by robotic solutions~\cite{zhou2022intelligent}.
The typical phases of selective harvesting are fruit detection and localization, end-effector motion planning, fruit attachment to the end-effector, fruit detachment from the plant, and transport to a storage container.
The surveys compiled over the years~\cite{bac2014harvesting, kootstra2021selective, zhou2022intelligent, rajendran2023towards} show that while substantial progress has been made in fruit detection and robotic hardware customization, the harvesting systems are still not ready for commercialization due to low success rates and high cycle times.

Whereas most attempts at autonomous harvesting have focused on citrus fruits or apples due to sparse foliage and easier localization, there have been only three reported attempts on the development of a full pipeline for sweet pepper harvesting: CROPS~\cite{bac2017performance}, Harvey~\cite{lehnert2020performance}, and SWEEPER~\cite{arad2020development}.
Sweet peppers are among the most difficult crops to autonomously harvest due to variation in shape and size, and severe occlusions by leaves leading to failures in both pepper and peduncle localization~\cite{bac2017performance}.

In CROPS~\cite{bac2017performance}, the focus of the research was on end-effector design, with color-based pepper detection and time of flight measurement for 3D localization.
Bac\etal\cite{bac2017performance} also developed a stem-dependent grasp pose calculation.
However, neither sweet pepper pose estimation nor peduncle localization was the focus of this work, which led to low success rates and high cycle times.

In Harvey~\cite{lehnert2020performance}, a sweet pepper pose estimation and grasping algorithm~\cite{lehnert2016sweet} together with MiniInception~\cite{mccool2017mixtures}, a mixture of lightweight CNN approach for peduncle segmentation, was deployed to improve the harvesting performance.
However, the peduncle localization accuracy is still limited with an F1-score of 0.502 and led to detachment failures.
Furthermore, Harvey used a customized end-effector with a suction cup and did not focus on motion planning for crop damage avoidance, or active perception.

Arad\etal\cite{arad2020development} focused on finding the best fit crop conditions and on testing\,\&\,validation of SWEEPER in a commercial glasshouse.
Semantic segmentation-based fruit and stem detection were deployed on a 6-DoF industrial robot arm with a customized end-effector, which caught the fruit after harvesting.
Due to the lack of peduncle localization, cutting failures were reported.

To the best of our knowledge, HortiBot is the first attempt at selective harvesting in general, and sweet peppers in particular, that focuses on all the aspects of harvesting: fruit detection and peduncle localization, active perception, environment-aware motion planning and force sensing-enabled adaptive manipulation.
HortiBot is a general-purpose system that can also be used for other horticulture operations such as leaf pruning and pollination.

\section{Hardware Setup and System Overview}
\label{sec:hardware}

While we focus on selective harvesting in this work, HortiBot is intended for autonomous operation of different horticulture operations such as leaf pruning, pollination, and crop monitoring.
This necessitates the use of dual-arm manipulation.
Additionally, an articulated head is necessary to enable the manipulation system to perceive in the presence of occlusions due to leaves and other plant organs during task completion.
The effectiveness of using a camera mounted on an arm for visual tele-manipulation has already been demonstrated~\cite{rakita2018autonomous, lenz2023nimbro}.

\subsection{Hardware Setup}

\begin{figure}[t]
	\centering
\tikzset{
    n/.style={fill=yellow!20,rounded corners,draw=black,text=black,align=center,thin,font=\footnotesize},
    lab/.style={draw=orange,latex-,thick}
 }
 \centering
 \tikz[]{

\def\nh{0.85}

\node[anchor=south west,inner sep=0] (image) at (0,0) {\includegraphics[width=0.9\columnwidth]{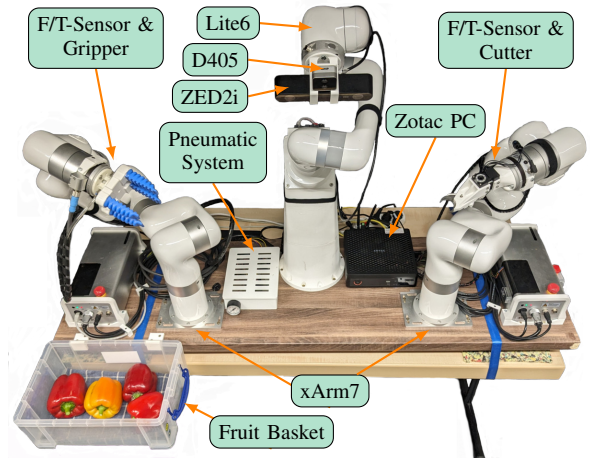}};
    \begin{scope}[x={(image.south east)},y={(image.north west)}]

        \draw[lab] (0.18,0.65) -- (0.15,\nh) node [n,manip,anchor=south] {F/T-Sensor \&\\Gripper};
        \draw[lab] (0.83,0.65) -- (0.85,\nh) node [n,manip,anchor=south] {F/T-Sensor \&\\Cutter};
        
        \draw[lab] (0.32,0.28) -- (0.55,0.11) node [n,manip,anchor=south] (xarm) {xArm7};
        \draw[lab] (0.72, 0.28) -- (xarm);
        
        \draw[lab] (0.52,0.94) -- (0.43,0.95) node [n,manip,anchor=east] {Lite6};
        \draw[lab] (0.54,0.85) -- (0.41,0.87) node [n,manip,anchor=east] {D405};
        \draw[lab] (0.48,0.81) -- (0.41,0.79) node [n,manip,anchor=east] {ZED2i};
        
        \draw[lab] (0.42,0.45) -- (0.35,0.6) node [n,manip,anchor=south] {Pneumatic\\System};
        
        \draw[lab] (0.3,0.12) -- (0.45,0.02) node [n,manip,anchor=south] {Fruit Basket};
        \draw[lab] (0.65,0.48) -- (0.73,0.70) node [n,manip,anchor=south] {Zotac PC};
    \end{scope}

}
 	\caption{HortiBot hardware setup.}
	\label{fig:system}
\end{figure}

HortiBot consists of two 7-DoF UFactory xArm7 and one 6-DoF UFactory Lite6 equipped with sensors and end-effectors to autonomously perform greenhouse applications such as selective harvesting in an adaptive manner (see \figref{fig:system}).
The system is mounted on the PATHoBot platform~\cite{smitt2021pathobot}, designed to operate in commercial glasshouse environments using the available structure.
It navigates on pipe-rails between individual crop rows and uses a scissor-lift to bring HortiBot to the desired height (up to 3\,m).

Both UFactory xArm7s are equipped with an OnRobot HEX-E force-torque sensor.
The right arm has a pneumatic four-finger soft gripper referred to as \textit{Grasper}.
The pneumatic pump and two air valves are controlled using the digital outputs of the xArm controller.
The left arm is equipped with a custom designed 1-DoF scissor, referred to as \textit{Cutter}.
The Lite6 carries two stereo cameras: a ZED2i stereo camera with deep learning based depth inferencing for medium range sensing, and a RealSense D405 for short range sensing, referred to as \textit{Observer}.
The Zed2i with a wide angle field of view of 110\,° has better performance in sunlight with its stereo based depth sensing and polarized lenses.

All three arms are connected to a common emergency-stop button for safe operation.
All necessary components including the control PC (Zotac ZBox with core i7-13700H, 32GB RAM and RTX4070 mobile GPU) running ROS-Noetic on Ubuntu 20.04, are mounted on a wooden platform which can be fitted onto the PATHoBot platform easily.

\subsection{System Overview}
\label{sec:sys_overview}
\begin{figure}[t]
	\centering
\begin{tikzpicture}[
 	font=\footnotesize,
    every node/.append style={text depth=.2ex},
	box/.style={rectangle, inner sep=0.05, anchor=west, align=center},
	line/.style={black, thick},
	midway/.append style={font=\footnotesize},
	above/.append style={yshift=-0.5ex},
	below/.append style={yshift=0.5ex},
	scale=0.8
]
\tikzset{every node/.append style={node distance=3.0cm}}
\tikzset{terminal_node/.append style={minimum size=1.0em,minimum height=1.7em,minimum width=1.5cm,draw,align=center,rounded corners,fill=yellow!40}}
\tikzset{l/.style={}}
\tikzset{dash_node/.append style={minimum size=1.5em,minimum height=1em,minimum width=2.4cm,align=left, rounded corners, font=\scriptsize}}
\tikzstyle{decision} = [diamond, minimum width=3cm, minimum height=1cm, text centered, draw=black, fill=green!30]

\def\x{2.7}
\def\y{2}

\node(perception)[terminal_node,perception] at(0,0) {Initial\\Mapping};
\node(select)[terminal_node,system] at(\x,0) {Select\\Fruit};
\node(fruit_follow)[terminal_node,perception] at(2*\x,0.6*\y) {Follow Fruit};
\node(approach)[terminal_node,manip] at(2*\x,0) {Approach\\Fruit};
\node(grasp)[terminal_node,otg] at(3*\x, 0) {Grasp Fruit};
\node(pull)[terminal_node,manip] at(3*\x,-\y) {Pull Fruit};
\node(cut)[terminal_node,otg] at(2*\x,-\y) {Cut Peduncle};
\node(place)[terminal_node,manip] at(1*\x,-\y) {Place Fruit};
\node(peduncle_follow)[terminal_node,perception] at(2*\x,-1.5*\y) {Follow Peduncle};

\draw [l,-latex] (perception.east) -- (select.west) node [midway,right] {};
\draw [l,-latex] (select.east) -- (approach.west) node [midway, above] {};
\draw [l,-latex] (approach.east) -- (grasp.west) node [midway,right] {};
\draw [l,-latex] (grasp.south) -- (pull.north) node [midway,right] {};
\draw [l,-latex] (pull.west) -- (cut.east) node [midway,right] {};
\draw [l,-latex] (cut.west) -- (place.east) node [midway,right] {};
\draw [l,-latex] (place.north) -- ++(0,0.3) -| node[left, pos=0.6, align=left]{If Fruits\\Remain} (select.south);
\draw [l,-latex] ([shift={(45:2pt)}]pull.south west) -- ([shift={(45:-2pt)}]peduncle_follow.north east) node [midway,right] {};
\draw [l,-latex] ([shift={(45:-2pt)}]select.north east) -- ([shift={(45:2pt)}]fruit_follow.south west) node [midway,right] {};

\end{tikzpicture} 
 	\caption{Workflow for autonomous selective pepper harvesting. Colors depict different actions: \compb[perception]{Perception} (see \secref{sec:perception}), \compb[system]{Logic} (see \secref{sec:auto_behaviour}), manipulation using \compb[manip]{Motion Primitives} (see \secref{sec:pmp}), and \compb[otg]{Online Trajectory Generation} (see \secref{sec:otg}).}
	\label{fig:behavior_tree}
\end{figure}
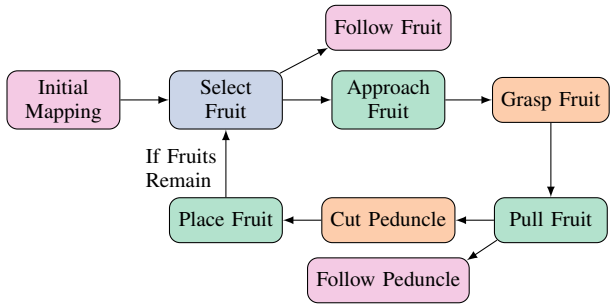

We use an adaptive autonomous behavior approach to perform selective harvesting.
\figref{fig:behavior_tree} shows a brief overview of the different phases of the autonomous harvesting workflow.
We carry out an initial mapping of the sweet peppers as described in \secref{sec:perception} to create a world model of the pepper plants.
Using this model, we select the fruits based on their reachability.
Thereafter, we activate the online fruit following while simultaneously approaching the fruit with the \textit{Grasper} and the \textit{Cutter}.
Once, we grasp the fruit, we pull it and then refine the \textit{Cutter} pose based on the updated peduncle localization from the \textit{Observer} as described in \secref{sec:manipulation}.
We adaptively adjust the cutter position based on force feedback and cut the peduncle after which the \textit{Grasper} transports it to the storage container and places it there.
Except for the initial mapping, the cycle is repeated until no fruits remain.

\subsection{Workspace Analysis}
\label{sec:workspace_analysis}
\begin{figure}[b]
	\centering
	\includegraphics[width=0.97\columnwidth,clip,trim=250px 0px 250px 400px]{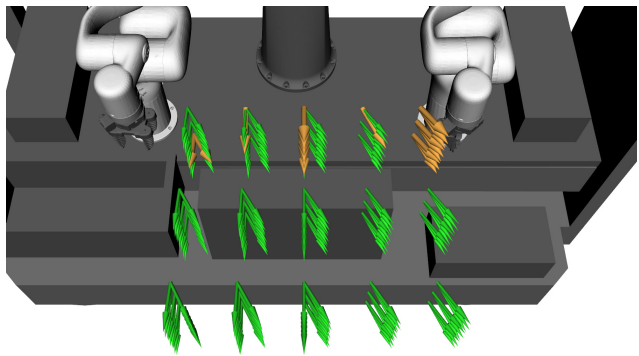}\vspace*{-1ex}
	\caption{Workspace analysis: Reachable (green) and non-reachable (yellow) fruit poses for both manipulation arms in the selected arm configuration.}
	\label{fig:workspace}
\end{figure}

The limited space in the glasshouse crop rows, and arm specifications~(workspace and kinematics) must be taken into account when designing the platform.
The goal is to maximize the common workspace between the two manipulation arms while reducing the potential for collisions.
We sampled 840~different arm mounting poses for both arms with different x- and y-positions and roll and pitch angles and tested each by counting collision-free IK-solutions reaching 270 sampled fruit poses.
For each sample, the grasp and cut end-effector pose is calculated for the corresponding arm.
The fruit poses have been sampled to be comparable to real poses in glasshouses and are shown in \figref{fig:workspace}.
Over 90\% of the sampled fruits are reachable with both arms in the selected configuration.
Repositioning the PATHoBot platform allows to access the remaining fruits.

\subsection{Calibration}
Some transformations must be calibrated before the system can be used with the required accuracy.
We perform a classical hand-eye calibration approach to estimate the transformations between \textit{Grasper's} and \textit{Cutter's} mounting poses, \textit{Grasper's} and \textit{Observer's} mounting poses and the camera mounting pose, similar to \cite{schwarz2021low}.
Custom 3D printed magnetic ArUco markers can be attached to the \textit{Grasper's} and \textit{Cutter's} last link with known transformations.
We collect about 2,000 samples with different arm configurations, extract pixel location in the images using the ArUco marker detection of the OpenCV library and computed the projected marker location into the image plane using forward-kinematics.
Finally we minimize the squared error function over all samples, yielding the optimized transformations with a mean reprojection error of 2.8\,pixels over the recorded samples.

In addition to the hand-eye calibration, the force-torque sensors need to be calibrated.
We collect 45 force-torque measurements from different sensor poses and use a standard least squares solver to determine the optimal parameters for the end-effector mass, the 3D center of mass with respect to the sensor frame, and the force and torque bias.
The calibration procedure is performed once after hardware changes or when the sensor bias drift becomes too large.
Currently, sensor drift is not compensated online, instead we use the relative change over a short time horizon.

\section{Perception and World Modeling}
\label{sec:perception}
\begin{figure}[t]
	\centering
\begin{tikzpicture}[
 	font=\footnotesize,
    every node/.append style={text depth=.2ex},
	box/.style={rectangle, inner sep=0.05, anchor=west, align=center},
	line/.style={black, thick},
	midway/.append style={font=\footnotesize},
	above/.append style={yshift=-0.5ex},
	below/.append style={yshift=0.5ex},
	scale=0.8
]
\tikzset{every node/.append style={node distance=3.0cm}}
\tikzset{terminal_node/.append style={minimum size=1.0em,minimum height=1.7em,minimum width=1.7cm,draw,align=center,rounded corners,fill=yellow!40}}
\tikzset{l/.style={}}

\def\x{3}
\def\y{1.7}

\node(s_detection)[terminal_node,system] at(\x,2*\y) {Stem\\Detection};
\node(stem_line)[terminal_node,system] at(2 *\x,2*\y) {Stemline\\Estimator};

\node(cam)[terminal_node, hardware] at (0, \y) {Zed2i\\Camera};
\node(p_detection)[terminal_node,manip] at(\x,\y) {Peduncle\\Detection};
\node(peduncle_loc)[terminal_node,manip] at (2*\x,\y){Peduncle\\Localization};

\node(f_detection) [terminal_node,manip] at(0,0) {Pepper\\Detection};
\node(mapping)[terminal_node,system] at(\x,0) {3D\\Mapping};
\node(shape_comp)[terminal_node,manip] at(2*\x,0) {Shape\\Completion};

\draw [l,-latex] (cam.south) -- (f_detection.north) node [midway,left] {\scriptsize RGB};
\draw [l,-latex] (cam.north) |- (s_detection.west) node [pos=0.2,left] {\scriptsize RGB};

\draw [l] ([xshift=0.1cm] cam.north) -- ++(0,0.35) -|  (2*\x,1.5*\y) node [pos=0.06,above] {\scriptsize Depth};
\draw [l,-latex] (2*\x,1.5*\y) -- (stem_line.south);
\draw [l,-latex] (2*\x,1.5*\y) -- (peduncle_loc.north);

\draw [l,-latex] (0.5*\x,1.5*\y) -- ++(down:1.4 * \y cm) |- ([yshift=0.1cm]mapping.west);

\draw [l,-latex] (s_detection.east) -- (stem_line.west);

\draw [l,-latex] (mapping.east) -- (shape_comp.west);
\draw [l,-latex] ([shift={(45:-2pt)}]f_detection.north east) -- ([shift={(45:2pt)}]p_detection.south west);
\draw [l,-latex] ([yshift=-0.1cm] f_detection.east) -- ([yshift=-0.1cm]mapping.west);
\draw [l,-latex] (p_detection.east) -- (peduncle_loc.west);

\draw [l,-latex] (stem_line.east) -- ++(0.5*\x,0) node [midway,above] {\scriptsize Stems};
\draw [l,-latex] (peduncle_loc.east) -- ++(0.5*\x,0) node [midway,above] {\scriptsize Peduncles};
\draw [l,-latex] (shape_comp.east) -- ++(0.5*\x,0) node [midway,above] {\scriptsize Fruits};

\end{tikzpicture}
	\caption{Perception pipeline \compb[hardware]{Hardware}, \compb[system]{Initial Mapping}, \compb[manip]{Common}. Stem detection and 3D mapping is done only during the initial mapping phase. Pepper and peduncle detection are carried out during initial mapping (\secref{sec:3d_mapping}) and fruit following (\secref{sec:fruit_following}). }
	\label{fig:perception_pipeline}
\end{figure}
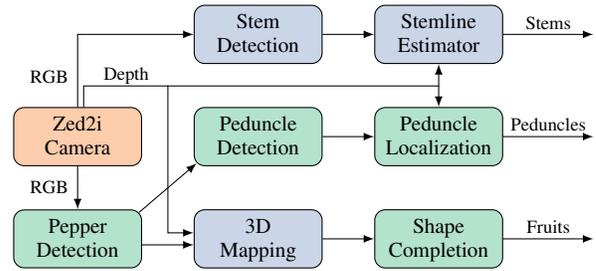

A dynamic world model of the pepper plants is necessary for successful autonomous harvesting.
To this end, perception of pepper plant organs is carried out in two stages as shown in \figref{fig:perception_pipeline}.
In the initial mapping stage, the manipulation arms are in stowed position and the \textit{Observer} records the pepper plant detections at different poses to create a world model of the pepper plants with sweet pepper fruits, associated peduncles and nearby stems.
The manipulation system uses this pepper plant model to determine the reachable fruits and selects them serially for harvesting.
During the manipulation phase, the \textit{Observer} performs online fruit following (\secref{sec:fruit_following}) for dynamic fruit and peduncle localization to account for perturbations in the fruit locations owing to the manipulation arms touching parts of the plants.
\subsection{Plant Organ Detection}
\label{sec:plant_organ_detection}
For creating a world model of pepper plants, we need to detect and localize the pepper fruits, peduncles and stems.
We adopted a multi-pronged approach for detecting these plant organs.
We combined the synthetic capsicum dataset~\cite{barth2016synthetic}, Kaggle sweet pepper dataset~\cite{montoya2021sweet}, and BUP20 dataset~\cite{halstead2020fruit}, to create an extensive dataset resulting in more than 130,000 instances of sweet pepper.
Since the synthetic dataset provides only semantic segmentation annotations, the detection of instances utilized OpenCV's~\cite{bradski2000opencv} contour finding and refinement to generate instance segmentation masks for sweet peppers and peduncles.

Reliable peduncle detection is necessary for the \textit{Cutter} to find the cutting point.
However, detecting peduncles in the full image is a challenging task as the mean Average Precision mAP@50 was only 0.435, for a model trained on the aforementioned dataset using the full images.
Hence, we developed a new approach for peduncle detection using cropped images.
From the original dataset, we created an additional dataset containing cropped images of size 96x96 pixels centered around the sweet peppers, annotated with pepper fruit and peduncle masks.
The cropped peduncle dataset contained more than 50,000 instances of peduncles and more than 100,000 instances of sweet peppers.
The results of the cropped peduncle detection method are presented in \secref{sec:ped_det}.

\begin{figure}[t]
	\centering
	\includegraphics[width=0.97\linewidth]
	{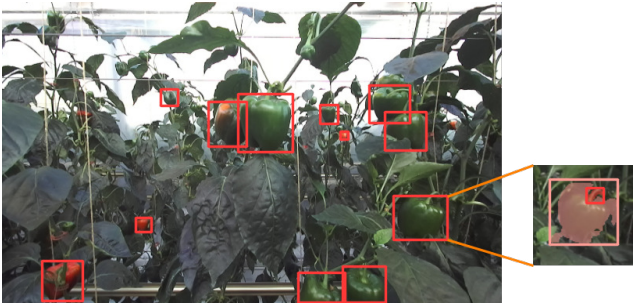}
	\caption{Cropped peduncle detection. Pepper and cropped peduncle detection applied on the fruits in the Campus Klein Altendorf glasshouse pepper plants. As can been seen, there are multiple peppers with peduncles not easily identifiable in the full image. The image on the right shows the cropped image obtained by inflating the pepper's bounding box and the resultant pepper and peduncle detected.}
	\label{fig:cropped_peduncle_detection}
\end{figure}
During run-time, the sweet pepper instance segmentation model is used to detect peppers in the full image.
For each pepper detected, a cropped image is created by inflating the bounding box by 50\,\%  as shown in \figref{fig:cropped_peduncle_detection}.
The cropped peduncle instance segmentation model is applied on this cropped image, and the peduncle mask and bounding box, if any detected, are transferred to the full image.
The peduncle detections are annotated with the associated fruit instance id for subsequent merging in the 3D domain.
YOLOv8's tracking mode was utilized to enable tracking of the sweet peppers and their associated peduncles across images.

During our initial trials, the \textit{Grasper} used to accidentally grasp the stems, especially when the peppers were located behind the stems.
Hence, it was imperative to detect and localize stems to enable the manipulation system to select a grasp that avoids grasping the stem during the fruit grasping.
We trained DeeplabV3Plus-Pytorch's~\cite{florian2017rethinking} semantic segmentation model on the synthetic capsicum dataset for semantic stem detections.
At run-time, using contour detections, the semantic masks of the stems were converted to a YOLOv8 consistent instance detection format.

\subsection{3D Mapping}
\label{sec:3d_mapping}
Instead of an iterative search, detect, and harvest approach for every pepper, which leads to higher cycle times, the \textit{Observer} performs an initial mapping of the pepper plants using a fixed number of poses that span across the reachable fruit locations (\secref{sec:workspace_analysis}).
This also enables the fruits to be viewed from different observations poses leading to better shape estimation.
At each observation pose, the depth segments and pepper masks are combined to form the instance id and semantic id annotated point cloud segments.
We adapted Voxblox++~\cite{grinvald2019volumetric} to integrate the pepper cloud segments based on YOLOv8's tracked instance id, as well as geometric overlap, to form an instance aware surface map of the sweet pepper fruits.

We also implemented an instance layer based extraction of point clouds from the Voxblox++ map to obtain the merged yet partial sweet pepper shapes formed after the initial mapping.
The \textit{Observer} then performs superellipsoid fitting~\cite{marangoz2022fruit} to estimate the completed shape, fruit pose and fruit dimensions from the partial sweet pepper shapes.
To mitigate the problem of over-segmentation in the 3D domain due to YOLOv8 losing the instance id tracking on account of occlusions, it then performs 3D overlap detection using the completed shapes' poses and dimensions.
When the overlap exceeds a certain threshold, the fruit with the smaller proportion of observed surface, computed as in our previous work~\cite{menon23nbv}, is discarded.

\subsection{Pepper Plant Modeling}
\label{sec:pepper_plant_modeling}
It is not sufficient to detect and localize peppers, peduncles and stems separately.
They must also be associated with each other to form a usable world model that the manipulation system can utilize.

Once the 3D mapping motion is completed, merging and association of the detections at each \textit{Observer} pose is performed to build the pepper plant model.
As peduncle and stem point cloud segments have a relatively low number of points, their integration using Voxblox++ is not reliable.
The \textit{Observer} identifies and tracks peduncles (fruit stalks) and stems concurrently, while it performs the next mapping motion.
It maintains a separate set of peduncles and stem segments to which the existing detections are added or merged.

Once the completed shapes are estimated, the \textit{Observer} attaches the peduncles to the completed pepper shapes.
It extracts each peduncle's bottom and top points as fruit point and stem point, after removing any outliers.
If the peduncle's bottom point is close enough($\leq$1\,cm) to the top center of the pepper, it associates the peduncle to the pepper.
If multiple peduncle segments belong to the same pepper, it merges them and recalculates the peduncle endpoints accordingly.

For stem localization, the \textit{Observer} rejects stems that are too short.
It then estimates the stem's 3D line using PCL's~\cite{rusu20113d} 3D RANSAC model.
New stem detections are compared with existing ones, and if they align well, they are merged in a greedy manner, and the 3D line parameters are recalculated.
If it finds a stem close to a pepper (within 5\,cm in the x-y plane), it considers the pepper to be attached to that stem.

The \textit{Observer} feeds the entire plant model consisting of the peppers with associated peduncles and stems to the manipulation system for selective fruit harvesting.

\subsection{Online Fruit Following and Pose Update}
\label{sec:fruit_following}
Once the manipulation system selects a fruit for harvesting (see~\secref{sec:sys_overview}), it transmits the selected fruit id to the \textit{Observer} for fruit following.
The online perception comprises two concurrent threads: one, running at 5\,Hz, computes the viewpose and sets goals for the online trajectory generation method detailed in \secref{sec:otg}.
The other, running at 10\,Hz, updates the fruit's pose estimate using instantaneous sweet pepper and peduncle detections for grasp and cut pose refinement.

During the grasping phase, the \textit{Observer} fixates on the selected fruit by moving to the corresponding viewpose.
The viewpose position $p_{vp}$ is calculated in the local trolley frame where the x axis is aligned along the platform length, y axis pointing towards the fruits and z axis vertically aligned.
We need the \textit{Observer} arm to be above the \textit{Cutter} i.e. $p^{z}_{vp}$ to allow it easy access for peduncle cutting.
At the same time, the \textit{Observer} arm needs $p^{x}_{vp}$ to be away from the vertical plane of the bases of the manipulation arms, whereas $p^{y}_{vp}$ needs to maintain at least 35\,cm from the fruit center for improved localization.
$p_{vp}$ is computed using the fruit center $p_{f}$ as follows:
\begin{eqnarray}
p^{x}_{vp} &= &0.8* (p^{x}_{f} - p^{x}_{h}) + p^{x}_{h} \\
p^{y}_{vp} &= &p^{y}_{f}-0.35 \\
p^{z}_{vp} &= &p^{z}_{f} + l^{z}_{f} + 0.15
\end{eqnarray}
where $l_f$ and $p_h$ represents the fruit bounding box dimensions, and the position of the base of the head arm respectively.
The normalized direction vector $dir_{vp}$ for the orientation of the viewpose is computed as follows:
\begin{equation}
	dir_{vp} = \frac{{p_f}- p_{vp}}{\lVert p_{f}- p_{vp} \rVert}
\end{equation}
$dir_{vp}$ and $p_{vp}$ are combined to form the SE3 viewpose.
During the cutting phase, the \textit{Observer} moves 5\,cm closer to the fruit while moving 2cm higher as well for peduncle fixation.

During the grasping phase, for the online pepper pose update, the \textit{Observer} gets the currently detected pepper cloud segments, smooths them using a moving least squares filter and subsequently performs shape estimation.
In this phase, we bias the fitting on the currently detected peppers to be closer to the initial fruit center and fruit dimensions, under the assumption that the initial mapping estimates are better due to the multi-view merging.

If the center of a currently detected pepper's completed shape is less than 1\,cm away from the initial mapping estimate, we greedily assign this detection and its completed shape to be the current estimate of the fruit.
However, if the center of the completed shape is more than 1\,cm but less than 3\,cm, then the fruit is added to a list of potential candidates for the current estimate and we choose the one nearest to the original estimate.
Using complementary filtering, we filter the current estimates for the fruit centers as follows:
\begin{equation}
{p^{\mathit{filt}}_{f}} = \alpha \cdot {p^{\mathit{curr}}_{f}} + (1-\alpha) \cdot {p^{\mathit{prev}}_{f}}
\end{equation}
where ${p^{\mathit{curr}}_{f}}$ and ${p^{\mathit{prev}}_{f}}$, ${p^{\mathit{filt}}_{f}}$, represent the current, previous and filtered estimates of the fruit center.

Once the fruit is grasped, the \textit{Observer} switches to online peduncle localization only.
The cropped peduncle detection method relies on the grasped pepper being detected in the full image.
However, due to occlusions by the gripper, the pepper detection fails frequently which leads to downstream failures in the peduncle detection.
Hence, the \textit{Observer} uses the gripper tool center point (TCP) pose to construct a 3D bounding box using the fruit center $p_f$ and fruit dimensions $l_f$ as follows:
\begin{alignat}{4}
x_{\mathit{bound}} & = && p^{x}_{f} \pm l^{x}_{f} \hspace{.6cm} & y_{\mathit{bound}} & = && p^{y}_{f} \pm l^{y}_{f}\\
z^{\mathit{bottom}}_{\mathit{bound}} & = && p^{z}_{f} \hspace{.3cm} & z^{\mathit{top}}_{\mathit{bound}} & = && p^{z}_{f} + l^{z}_{f} + 0.05
\end{alignat}
The 3D bounding box points are converted to 2D points on the image using the camera parameters.
The region of interest for peduncle detection is computed as the minimum and maximum of the 2D points.
The peduncle detected in the RoI is converted to cloud segment and added to a buffer with length 4 and maximum age of 0.5\,s.
The valid frames of the buffer are merged and smoothed to obtain the updated fruit and stem points.

\section{Dual-Arm Manipulation}
\label{sec:manipulation}

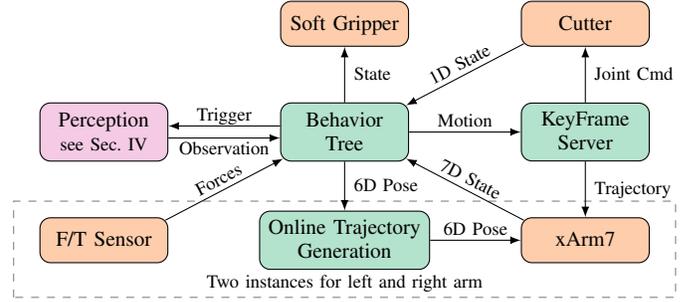
\begin{figure}
\centering
\begin{tikzpicture}[
 	font=\footnotesize,
    every node/.append style={text depth=.2ex},
	box/.style={rectangle, inner sep=0.05, anchor=west, align=center},
	line/.style={black, thick},
	midway/.append style={font=\footnotesize},
	above/.append style={yshift=-0.5ex},
	below/.append style={yshift=0.5ex},
	scale=0.8
]
\tikzset{every node/.append style={node distance=3.0cm}}
\tikzset{terminal_node/.append style={minimum size=1.0em,minimum height=1.7em,minimum width=1.7cm,draw,align=center,rounded corners,fill=yellow!40}}
\tikzset{l/.style={}}
\tikzset{dash_node/.append style={minimum size=1.5em,minimum height=1em,minimum width=2.4cm,align=left, rounded corners, font=\scriptsize}}

\def\x{4}
\def\y{1.8}

\node(gripper)[terminal_node,hardware] at(\x,2*\y) {Soft Gripper};
\node(xarm)[terminal_node,hardware] at(2*\x,0) {xArm7};
\node(cutter) [terminal_node,hardware] at(2*\x,2*\y) {Cutter};
\node(ft_sensor)[terminal_node,hardware] at(0,0) {F/T Sensor};

\node(otg)[terminal_node,manip] at(\x,0) {Online Trajectory\\Generation};
\node(keyframe)[terminal_node,manip] at (2*\x,\y){KeyFrame\\Server};

\node(bt)[terminal_node,manip] at(\x,\y) {Behavior\\Tree};

\node(perception)[terminal_node,perception] at(0,\y) {Perception\\\scriptsize{see \secref{sec:perception}}};

\def\dashsepx{1.5}
\def\dashsepy{0.15}
\def\nodeheight{0.8}

\draw [dashed, gray] (-\dashsepx, -\nodeheight-\dashsepy) rectangle (2*\x+\dashsepx, \nodeheight-\dashsepy);

\draw [l,-latex] (bt.north) -- (gripper.south) node [midway,right] {\scriptsize State};
\draw [l,-latex] (bt.east) -- (keyframe.west) node [midway,above] {\scriptsize Motion};
\draw [l,-latex] (bt.south) -- (otg.north) node [midway,right] {\scriptsize 6D Pose};
\draw [l,latex-] ([shift={(45:2pt)}]bt.south west) -- ([shift={(45:-2pt)}]ft_sensor.north east) node [midway,above,sloped] {\scriptsize Forces};
\draw [l,latex-] ([shift={(45:-2pt)}]bt.north east) -- ([shift={(45:2pt)}]cutter.south west) node [midway,above,sloped] {\scriptsize 1D State};
\draw [l,latex-] ([shift={(135:2pt)}]bt.south east) -- ([shift={(135:-2pt)}]xarm.north west) node [midway,above,sloped] {\scriptsize 7D State};

\draw [l,-latex] (otg.east) -- (xarm.west) node [midway,above] {\scriptsize 6D Pose};
\draw [l,-latex] (keyframe.north) -- (cutter.south) node [midway,right] {\scriptsize Joint Cmd};
\draw [l,-latex] (keyframe.south) -- (xarm.north) node [midway,right] {\scriptsize Trajectory};

\node()[dash_node] at (\x, -0.7) {Two instances for left and right arm};

\draw [l,latex-] ([yshift=0.1cm]perception.east) -- ([yshift=0.1cm]bt.west) node [midway,above] {\scriptsize Trigger};
\draw [l,-latex] ([yshift=-0.1cm]perception.east) -- ([yshift=-0.1cm]bt.west) node [midway,below] {\scriptsize Observation};

\end{tikzpicture} 
 \vspace{-2ex}
\caption{Manipulation hard- and software pipeline. Behavior tree-based decisions control both xArm7s using motion primitives or online trajectory generation based on perception observations and sensor measurements. Colors correspond to categories \compb[hardware]{Hardware}, \compb[manip]{Manipulation}, and \compb[perception]{Perception}.}
\label{fig:manipulation}
\end{figure}

Autonomous horticultural operations require adaptive manipulation capabilities to robustly handle different plant arrangements and cope with dynamic changes such as moving a fruit while grasping it.

Our dual-arm manipulation method controls two xArm7s, a soft gripper, and a custom cutter based on observations provided by the perception pipeline (see \secref{sec:perception}) and force-torque sensors attached to each arm (see \figref{fig:manipulation}).
The required motions vary in execution length and the goal pose update frequency, i.e.
long motions ($>$0.5\,sec) with a goal pose known before motion execution (for example approaching the fruit) and motions with a dynamic goal poses such as grasping and cutting the fruit, which have a non-zero start velocity.
We use two different motion generation and execution methods, \textit{Parameterized Motion Primitives (PMP)} and \textit{Online Motion Generation (OTG)} to handle these requirements, which are described in the following.

\begin{figure}[t]
	\centering
\tikzset{
    n/.style={fill=yellow!20,rounded corners,draw=black,text=black,align=center,thin,font=\footnotesize},
    lab/.style={draw=red,latex-,thick}
 }
 \centering
 \tikz[]{
  \node[inner sep=0pt] (img) {\includegraphics[width=0.97\columnwidth,clip,trim=350px 0px 350px 300px]{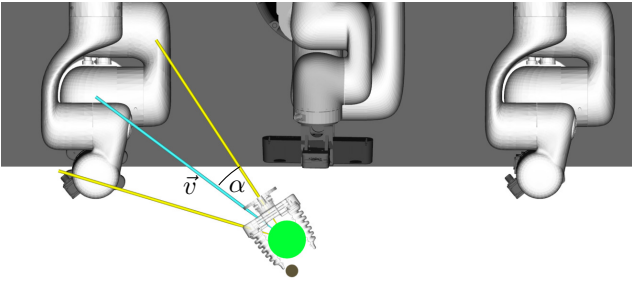}};

  \node at (-1.68,-0.58){$\vec{v}$};
  
  \node at (-1.05,-0.6){$\alpha$};
  
  \def\r{1.2}
  
  \draw (-0.38,-1.36) ++(140:\r) arc(140:122:\r);

}
 	\vspace{-1ex}
	\caption{Grasp direction (shown by the gripper model) for the fruit (green circle) is selected to be opposite of the corresponding stem (brown circle) without exceeding an angular deviation (yellow lines) from the vector $v$ connecting the arm mounting point and the fruit center (light blue line).
	If no corresponding stems are detected, $v$ is used as the grasp direction.}
	\label{fig:mani_grasp}
\end{figure}

\subsection{Parameterized Motion Primitives (PMP)}
\label{sec:pmp}
All motions with a fixed goal pose (specified offline or online), are generated using \textit{Parameterized Motion Primitives (PMP)}.
A PMP consist of one or multiple keyframes each specifying one or multiple kinematic chains to be manipulated.
The target configuration can be defined in Cartesian or joint space per chain and the generated motion linearly interpolates between the start and each keyframe goal configuration.
We use \textit{nimbro$\_$ik}~\cite{schwarz2017nimbro} to generate joint space configurations for Cartesian goal poses which uses a \textit{selectively damped least squares (SDLS)} solver~\cite{buss2005selectively} and allows to define cost-functions which are optimized in the null-space.
In this setup, we penalize the elbow crossing a vertical plane towards the system's center to reduce potential collisions.
Self-collisions are checked along the trajectory and reported before motion execution.
PMPs are either predefined offline for static motions such as placing the fruit in the container, or are parametrized online using sensor data for example when approaching the fruit.

\subsection{Online Trajectory Generation (OTG)}
\label{sec:otg}
Since every trajectory generated using PMP assumes zero start velocity, online replanning is not feasible.
Instead, we switch the xArm control mode to OTG, which allows online replanning to follow a 6D end-effector goal pose considering the current robot state including current joint velocities and velocity and acceleration limits.
However, this control mode does not provide any kind of (self-) collision checks.
Therefore, we added collision checking on top of the OTG control mode.
We compute the minimal distance $dist_{c}$ between any two links at the current robot state and $dist_{n}$ for an extrapolated state assuming constant velocity using \textit{MoveIt}~\cite{coleman2014reducing}.
Next, we compute $\alpha_{\{c,n\}}$ which is used to reduce the motion velocity when approaching a collision:

\begin{equation}
\alpha_{\{c,n\}} = \frac{dist_{\{c,n\}} - 0.5}{3.0 - 0.5}
\end{equation}

and ensure that, $\alpha_{\{c,n\}} \in [0,1]$.
If $alpha_{n} \le alpha_{c}$, i.e., the robot is approaching a self-collision, we reduce the current motion velocity exponentially with the scalar $\beta \in [0,1]$ which is defined as follows:

\begin{equation}
\beta = \frac{2^{5 \alpha_{n}} - 1}{2^{5}-1}
\end{equation}

This prevents the arm from self-colliding and but does not prevent the arm to move away from collisions.

In addition, the OTG controller reports the current status.
The control-loop runs with 30\,Hz which is sufficient for our application.
We run three instances of this controller, one for each xArm7, and the \textit{Observer} in case of online fruit following (see \secref{sec:fruit_following}).

\subsection{Adaptive Manipulation}
\label{sec:auto_behaviour}

\begin{figure}[t]
	\centering
\tikzset{
    n/.style={fill=yellow!20,rounded corners,draw=black,text=black,align=center,thin,font=\footnotesize},
    lab/.style={draw=orange,latex-,thick}
 }
 \centering
 \tikz[]{
  \node[anchor=south west,inner sep=0] (img) at (0,0) {\includegraphics[height=4.0cm,clip,trim=900px 10px 00px 300px]{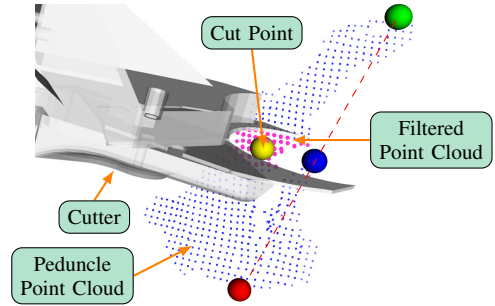}};
    \begin{scope}[x={(img.south east)},y={(img.north west)}]

        \draw[lab] (0.2,0.45) -- (0.15,0.35) node [n,manip,anchor=north] {Cutter};
        \draw[lab] (0.58,0.52) -- (0.55,0.85) node [n,manip,anchor=south] {Cut Point};
    
        \draw[lab] (0.4,0.2) -- (0.25,0.1) node [n,manip,anchor=east] {Peduncle\\Point Cloud};
        \draw[lab] (0.65,0.55) -- (0.85,0.55) node [n,manip,anchor=west] {Filtered\\Point Cloud};

        \draw[red, dashed] (0.915,0.945) -- (0.52,0.045);

    \end{scope}
}
	\caption{Cut pose (shown by the cutter model) is computed using the peduncle point cloud. Green and red spheres indicate the highest and lowest peduncle points. Magenta points show the filtered peduncle cloud using a box filter centered at the blue sphere. The cut position (yellow sphere) is the centroid of the filtered cloud.}
	\label{fig:mani_cut}
\end{figure}

Autonomous selective and adaptive harvesting requires a system which acts based on various sensor measurements.
We calculate grasp and cut poses based on the online perception results and adjust manipulation trajectories using force-torque measurements.

The grasp direction is always orthogonal to the fruit's main axis (which is mostly vertical).
The grasp direction is computed based on the fruit position and stem detection (see \figref{fig:mani_grasp}).
Let $\vec{v}\in R^2$ be the vector between the \textit{Gripper's} mounting center and the fruit center projected onto the x-y plane.
We use $\vec{v}$ as the grasping direction if no stem detection is available for the selected fruit.

If stem detections are available, the fruit is grasped such, that the stem is avoided as much as possible.
Let $\beta$ be the angle between $\vec{v}$ and the selected grasp direction.
We select the grasp direction opposite of the stem, with the condition $\beta \le \ang{20}$.
This avoids the stem (as much as possible) while creating feasible arm configurations.

We use the peduncle point cloud $PC$ to calculate the optimal cut pose (see \figref{fig:mani_cut}).
Let $\vec{p}$ be the vector connecting the highest and lowest point of $PC$ (in the vertical axis) and let $M$ be the midpoint of $\vec{p}$.
We filter $PC$ using a box filter which is centered at $M$ and aligned in the direction of $\vec{p}$.
The centroid of the filtered $PC$ is used as the cutting position.
The cutting orientation is fixed relative to the \textit{Cutter's} mounting pose, similar to the grasp orientation.
We use a fixed cut position above the fruit in case of missing peduncle detections and ensure a minimum distance between grasp and cut pose to avoid self-collisions.

We update the grasp and cut pose  with 100\,Hz using the latest perception results and generate new arm trajectories using OTG (see \secref{sec:otg}).
In addition, we detect reaching the fruit or peduncle while grasping and cutting by monitoring the force measurements relative to the start of the motions.
The current motion is stopped if the observed forces exceed a predefined threshold.

\section{Results}
\label{sec:exp}

\begin{figure}[t]
	\centering
	\includegraphics[width=\linewidth]{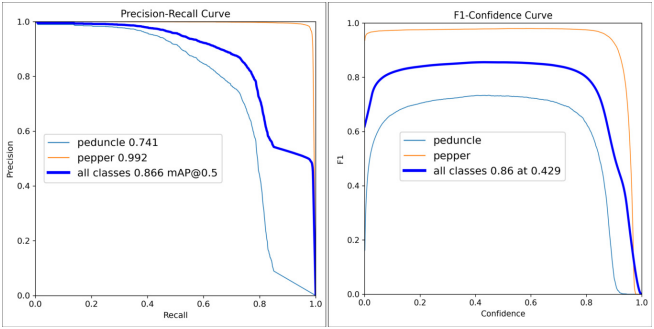}
	\caption{Mask PR curve and F1 curve for peduncle detection in cropped image.}
	\label{fig:mask_pr_curve}
\end{figure}

We evaluated two major aspects of our approach, namely peduncle detection and the adaptive autonomous selective harvesting performance.
In the latter, we evaluate the success rate and execution time for different phases of the approach.
\subsection{Peduncle Detection}
\label{sec:ped_det}
While sweet pepper detection is a fairly mature research area, peduncle detection still has a lot of scope for improvement, with the MiniInception~\cite{mccool2017mixtures} approach reporting an F1 score of 0.313 and 0.564 for unfiltered and filtered data, respectively.
Our method demonstrates a far superior performance with a mean average precision mAP@50 of 0.741 for peduncle segmentation as can be seen in \figref{fig:mask_pr_curve}. Similarly, the F1 score for peduncle detection is significantly better at 0.781.

The mean processing time for both pepper and peduncle detection and localization is 100 ms, which enables the system to perform these tasks in real time.

\subsection{Experimental Harvesting Setup}
\label{sec:exp_setup}

\begin{figure}[t]
 \centering
 \includegraphics[width=0.7\linewidth,clip,trim=0px 1600px 500px 400px]{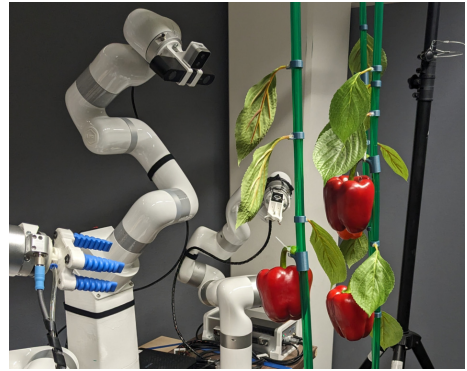}
 \caption{Experimental harvesting setup.}
 \label{fig:harvesting_setup}
\end{figure}

We created an indoor mock-up of sweet pepper plants using real sweet peppers, artificial leaves and thin pipes as stems (see \figref{fig:harvesting_setup}).
3D printed holders were used to attach the leaves and peduncles to the stems.
We conducted six trials with 4 fruits to be harvested in every trial.
The fruits were distributed among 3 stems.
We used red (21), yellow (1) and orange (2) sweet peppers with sufficiently long peduncle for mounting reasons.
The fruits were located roughly 0.5\,m away from the HortiBot platform, similar to the glasshouse rows.
The initial mapping was carried out using 5 observation poses.
We did not use the D405 camera as it is needed only for close range sensing in the glasshouse.

\subsection{Harvesting Trials Results}
\label{sec:exp_eval}

\begin{table}
	\centering
	\footnotesize
	\caption{Experimental Results.}\label{tab:exp_success}
	\begin{tabular}{lrrrrr}
		\toprule
		Trial & \multicolumn{4}{c}{Success} & Time\\
		\cmidrule (lr) {2-5}
		& Grasp & Cut & Place & Overall & [m:s]\\
		\midrule
		1 & 4/4 & 3/4 & 3/3 & 3/4 & 1:45 \\
		2 & 4/4 & 3/4 & 3/3 & 3/4 & 1:47 \\
		3 & 4/4 & 3/4 & 3/3 & 3/4 & 1:49 \\
		4 & 4/4 & 4/4 & 3/4 & 3/4 & 1:49 \\
		5 & 4/4 & 4/4 & 4/4 & 4/4 & 1:49 \\
		6 & 4/4 & 4/4 & 4/4 & 4/4 & 1:48 \\
		\midrule
		Total & 24/24 & 21/24 & 20/21 & 20/24 &\\
		\bottomrule
	\end{tabular}
\end{table}
We evaluated the full system pipeline by analyzing the execution time (recorded automatically) and the success rate (recorded manually) for each phase.\footnote{\scriptsize \url{https://www.ais.uni-bonn.de/videos/IROS_2024_Lenz/}}
The overall success rate for the entire harvesting cycle is 83.33\% with 20 out of a total of 24 fruits harvested successfully, as can been seen in \tabref{tab:exp_success}.
HortiBot was able to successfully grasp all fruits without using any expensive grasp detection approach.
The flexible pneumatic gripper perfectly adapted to the fruit shape.
This validates our approach to use shape completion based grasp pose estimation with force sensing enabled compliant grasping.
The cutting phase had a lower success rate of 87.5\% with 21 out of 24 peduncles cut.
The three failures were due to peduncle localization errors with depth registration issues.
While peduncle detection in the RGB image was successful in all the cases, the depth rendering was poor due to the thin structures.
While transporting the fruit, we had one failure in trial 4 due to an imperfect grasp resulting in the fruit slipping out of the gripper.

\pgfplotstableread[row sep=\\,col sep=&]{ state & mean & se \\ Mapping &11.98 & 0.19 \\ Approach & 3.54 & 0.28 \\ Grasp &4.76 & 0.11\\ Cut & 10.95 & 0.17\\ Place & 4.7 & 0.08\\ }\mydata
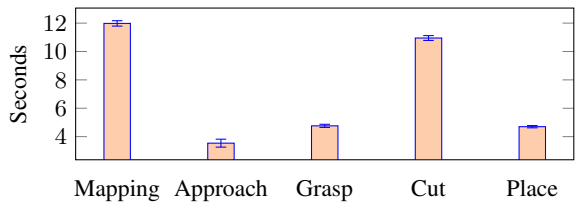
\begin{figure}[t]
	\centering
	\small
	\begin{tikzpicture}
	\begin{axis}[
	ybar,
	height=3.6cm,
	width=.95\linewidth,
	symbolic x coords={Mapping, Approach, Grasp, Cut, Place},
	xtick=data,
	ylabel={Seconds},
	xtick style={draw=none},
	ytick={2,4,...,12},
	]
	\addplot+ [
	hardware,
	mark options={red, scale=2.75},
	error bars/.cd,
	y fixed,
	y dir=both,
	y explicit
	] table[x=state,y=mean,y error=se]{\mydata};
	\end{axis}
	\end{tikzpicture}\vspace*{-2ex}
	\caption{Mean execution time and standard error for different phases of the harvesting cycle over six trials with four fruits each. Note, mapping is performed once per trial, all other phases once per fruit.}
	\label{fig:exp_times}
\end{figure}
The execution time for harvesting is another key factor for determining the performance of the system.
The overall mean time needed for each trial was 1\,min 48\,s, thus leading to an average time of 26.95\,s per fruit including the failure cases.
As can been seen in \figref{fig:exp_times}, the initial mapping needed 11.98\,s on an average for a total of 5 poses with all the fruits successfully detected in all the trials.
Using \textit{PMP}, the approach phase required an average of only 3.54\,s per fruit for the \textit{Grasper} and the \textit{Cutter} to reach their respective pre-grasp and pre-cut poses.
The grasp phase required 4.76\,s per fruit including opening  and closing the gripper, and pre-grasp to grasp pose \textit{OTG} motion.
However, it was the peduncle cutting that required the longest time with each fruit needing around 10.95\,s.
This was due to the noisy peduncle localization update which caused the \textit{OTG} cut motion to continue refining the cut pose until it found a stable pose.
The transport of fruits was relatively fast with the place phase needing only 4.7\,s per fruit.

\section{Summary}
\label{sec:concl}
In this work, we presented HortiBot, a fully integrated system that focuses on all aspects of robotic harvesting.
With an articulated head for active perception, and force-sensing enabled bi-manual manipulation, HortiBot can carry out different horticulture tasks.
We developed a novel peduncle detection method that has significantly better detection accuracy, leading to 87.5\% success in peduncle cutting.
We also developed a novel collision-aware online trajectory generation method that is able to perform pose tracking at 30\,Hz frequency.
Using force-sensing based compliant grasping and cutting, we achieved an overall success rate of 83.33\% and a cycle time of 27\,s per fruit on an indoor mock-up of pepper plants, which outperforms state-of-the-art selective harvesting robots.
We plan to deploy HortiBot mounted on the PATHoBot~\cite{smitt2021pathobot} for sweet pepper harvesting in glasshouse scenarios in the future.

\bibliographystyle{IEEEtran}

\balance
\bibliography{refs}

\begin{thebibliography}{10}
\providecommand{\url}[1]{#1}
\csname url@rmstyle\endcsname
\providecommand{\newblock}{\relax}
\providecommand{\bibinfo}[2]{#2}
\providecommand\BIBentrySTDinterwordspacing{\spaceskip=0pt\relax}
\providecommand\BIBentryALTinterwordstretchfactor{4}
\providecommand\BIBentryALTinterwordspacing{\spaceskip=\fontdimen2\font plus
\BIBentryALTinterwordstretchfactor\fontdimen3\font minus
  \fontdimen4\font\relax}
\providecommand\BIBforeignlanguage[2]{{%
\expandafter\ifx\csname l@#1\endcsname\relax
\typeout{** WARNING: IEEEtran.bst: No hyphenation pattern has been}%
\typeout{** loaded for the language `#1'. Using the pattern for}%
\typeout{** the default language instead.}%
\else
\language=\csname l@#1\endcsname
\fi
#2}}

\bibitem{bac2014harvesting}
C.~W. Bac, E.~J. Van~Henten, J.~Hemming, and Y.~Edan, ``Harvesting robots for
  high-value crops: State-of-the-art review and challenges ahead,''
  \emph{Journal of Field Robotics (JFR)}, vol.~31, no.~6, 2014.

\bibitem{kootstra2021selective}
G.~Kootstra, X.~Wang, P.~M. Blok, J.~Hemming, and E.~Van~Henten, ``Selective
  harvesting robotics: current research, trends, and future directions,''
  \emph{Current Robotics Reports}, vol.~2, pp. 95--104, 2021.

\bibitem{zhou2022intelligent}
H.~Zhou, X.~Wang, W.~Au, H.~Kang, and C.~Chen, ``Intelligent robots for fruit
  harvesting: Recent developments and future challenges,'' \emph{Precision
  Agriculture}, vol.~23, no.~5, 2022.

\bibitem{rajendran2023towards}
V.~Rajendran, B.~Debnath, S.~Mghames, W.~Mandil, S.~Parsa, S.~Parsons, and
  A.~Ghalamzan-E, ``Towards autonomous selective harvesting: A review of robot
  perception, robot design, motion planning and control,'' \emph{Journal of
  Field Robotics (JFR)}, 2023.

\bibitem{tong2024advancements}
Y.~Tong, H.~Liu, and Z.~Zhang, ``Advancements in humanoid robots: A
  comprehensive review and future prospects,'' \emph{IEEE/CAA Journal of
  Automatica Sinica}, vol.~11, no.~2, 2024.

\bibitem{li2022design}
K.~Li, Y.~Huo, Y.~Liu, Y.~Shi, Z.~He, and Y.~Cui, ``Design of a lightweight
  robotic arm for kiwifruit pollination,'' \emph{Computers and Electronics in
  Agriculture}, vol. 198, 2022.

\bibitem{you2023semiautonomous}
A.~You, N.~Parayil, J.~G. Krishna, U.~Bhattarai, R.~Sapkota, D.~Ahmed,
  M.~Whiting, M.~Karkee, C.~M. Grimm, and J.~R. Davidson, ``Semiautonomous
  precision pruning of upright fruiting offshoot orchard systems: An integrated
  approach,'' \emph{IEEE Robotics and Automation Magazine (RAM)}, 2023.

\bibitem{bac2017performance}
C.~W. Bac, J.~Hemming, B.~Van~Tuijl, R.~Barth, E.~Wais, and E.~J. van Henten,
  ``Performance evaluation of a harvesting robot for sweet pepper,''
  \emph{Journal of Field Robotics (JFR)}, vol.~34, no.~6, 2017.

\bibitem{lehnert2020performance}
C.~Lehnert, C.~McCool, I.~Sa, and T.~Perez, ``Performance improvements of a
  sweet pepper harvesting robot in protected cropping environments,''
  \emph{Journal of Field Robotics (JFR)}, vol.~37, no.~7, 2020.

\bibitem{arad2020development}
B.~Arad, J.~Balendonck, R.~Barth, O.~Ben-Shahar, Y.~Edan, T.~Hellstr{\"o}m,
  J.~Hemming, P.~Kurtser, O.~Ringdahl, T.~Tielen, \emph{et~al.}, ``Development
  of a sweet pepper harvesting robot,'' \emph{Journal of Field Robotics (JFR)},
  vol.~37, no.~6, 2020.

\bibitem{lehnert2016sweet}
C.~Lehnert, I.~Sa, C.~McCool, B.~Upcroft, and T.~Perez, ``Sweet pepper pose
  detection and grasping for automated crop harvesting,'' in \emph{IEEE
  Intl.~Conf.~on Robotics \& Automation (ICRA)}.\hskip 1em plus 0.5em minus
  0.4em\relax IEEE, 2016.

\bibitem{mccool2017mixtures}
C.~McCool, T.~Perez, and B.~Upcroft, ``Mixtures of lightweight deep
  convolutional neural networks: Applied to agricultural robotics,'' \emph{IEEE
  Robotics and Automation Letters (RA-L)}, vol.~2, no.~3, 2017.

\bibitem{rakita2018autonomous}
D.~Rakita, B.~Mutlu, and M.~Gleicher, ``An autonomous dynamic camera method for
  effective remote teleoperation,'' in \emph{ACM/IEEE Intl.~Conf.~on
  Human-Robot Interaction (HRI)}, 2018.

\bibitem{lenz2023nimbro}
C.~Lenz, M.~Schwarz, A.~Rochow, B.~P{\"a}tzold, R.~Memmesheimer, M.~Schreiber,
  and S.~Behnke, ``{NimbRo} wins ana avatar xprize immersive telepresence
  competition: Human-centric evaluation and lessons learned,''
  \emph{International Journal of Social Robotics}, 2023.

\bibitem{smitt2021pathobot}
C.~Smitt, M.~Halstead, T.~Zaenker, M.~Bennewitz, and C.~McCool, ``Pathobot: A
  robot for glasshouse crop phenotyping and intervention,'' in \emph{IEEE
  Intl.~Conf.~on Robotics \& Automation (ICRA)}.\hskip 1em plus 0.5em minus
  0.4em\relax IEEE, 2021.

\bibitem{schwarz2021low}
M.~Schwarz and S.~Behnke, ``Low-latency immersive 6d televisualization with
  spherical rendering,'' in \emph{IEEE-RAS Intl.~Conf.~on Humanoid
  Robots}.\hskip 1em plus 0.5em minus 0.4em\relax IEEE, 2021.

\bibitem{barth2016synthetic}
R.~Barth, ``Synthetic and empirical capsicum annuum image dataset,'' 2016.

\bibitem{montoya2021sweet}
L.~E. Montoya~Cavero, ``Sweet pepper recognition and peduncle pose
  estimation,'' 2021.

\bibitem{halstead2020fruit}
M.~Halstead, S.~Denman, F.~Clinton, and C.~McCool, ``Fruit detection in the
  wild: The impact of varying conditions and cultivar,'' in \emph{Digital Image
  Computing: Techniques and Applications (DICTA)}, 2020.

\bibitem{bradski2000opencv}
G.~Bradski, A.~Kaehler, \emph{et~al.}, ``Opencv,'' \emph{Dr. Dobb’s journal
  of software tools}, vol.~3, no.~2, 2000.

\bibitem{florian2017rethinking}
L.-C. Florian and S.~H. Adam, ``Rethinking atrous convolution for semantic
  image segmentation,'' in \emph{IEEE Conf.~on Computer Vision and Pattern
  Recognition (CVPR)}, vol.~6, 2017.

\bibitem{grinvald2019volumetric}
M.~{Grinvald}, F.~{Furrer}, T.~{Novkovic}, J.~J. {Chung}, C.~{Cadena},
  R.~{Siegwart}, and J.~{Nieto}, ``{Volumetric Instance-Aware Semantic Mapping
  and 3D Object Discovery},'' \emph{IEEE Robotics and Automation Letters
  (RA-L)}, vol.~4, no.~3, July 2019.

\bibitem{marangoz2022fruit}
S.~Marangoz, T.~Zaenker, R.~Menon, and M.~Bennewitz, ``Fruit mapping with shape
  completion for autonomous crop monitoring,'' in \emph{IEEE Intl.~Conf.~on
  Automation Science and Engineering (CASE)}.\hskip 1em plus 0.5em minus
  0.4em\relax IEEE, 2022.

\bibitem{menon23nbv}
R.~Menon, T.~Zaenker, N.~Dengler, and M.~Bennewitz, ``{NBV-SC}: Next best view
  planning based on shape completion for fruit mapping and reconstruction,'' in
  \emph{IEEE/RSJ Intl.~Conf.~on Intelligent Robots and Systems (IROS)}, 2023.

\bibitem{rusu20113d}
R.~B. Rusu and S.~Cousins, ``{3D} is here: Point cloud library {(PCL)},'' in
  \emph{IEEE Intl.~Conf.~on Robotics \& Automation (ICRA)}.\hskip 1em plus
  0.5em minus 0.4em\relax IEEE, 2011.

\bibitem{schwarz2017nimbro}
M.~Schwarz, A.~Milan, C.~Lenz, A.~Munoz, A.~S. Periyasamy, M.~Schreiber,
  S.~Sch{\"u}ller, and S.~Behnke, ``{NimbRo} picking: Versatile part handling
  for warehouse automation,'' in \emph{IEEE Intl.~Conf.~on Robotics \&
  Automation (ICRA)}.\hskip 1em plus 0.5em minus 0.4em\relax IEEE, 2017.

\bibitem{buss2005selectively}
S.~R. Buss and J.-S. Kim, ``Selectively damped least squares for inverse
  kinematics,'' \emph{Journal of Graphics tools}, vol.~10, no.~3, 2005.

\bibitem{coleman2014reducing}
D.~Coleman, I.~Sucan, S.~Chitta, and N.~Correll, ``Reducing the barrier to
  entry of complex robotic software: a moveit! case study,'' \emph{Journal of
  Software Engineering for Robotics}, 2014.

\end{thebibliography}
\balance

\end{document}